# A Differential Attention Fusion Model Based on Transformer for Time Series Forecasting

Benhan Li
*School of Mathematics*
*Southwest Jiaotong University*
Chengdu, China
bhli@my.swjtu.edu.cn

Shengdong Du*
*School of Computing and Artificial Intelligence*
*Southwest Jiaotong University*
Chengdu, China
sddu@swjtu.edu.cn

Tianrui Li
*School of Computing and Artificial Intelligence*
*Southwest Jiaotong University*
Chengdu, China
trli@swjtu.edu.cn

*Abstract*—Time series forecasting is widely used in the fields of equipment life cycle forecasting, weather forecasting, traffic flow forecasting, and other fields. Recently, some scholars have tried to apply Transformer to time series forecasting because of its powerful parallel training ability. However, the existing Transformer methods do not pay enough attention to the small time segments that play a decisive role in prediction, making it insensitive to small changes that affect the trend of time series, and it is difficult to effectively learn continuous time-dependent features. To solve this problem, we propose a differential attention fusion model based on Transformer, which designs the differential layer, neighbor attention, sliding fusion mechanism, and residual layer on the basis of classical Transformer architecture. Specifically, the differences of adjacent time points are extracted and focused by difference and neighbor attention. The sliding fusion mechanism fuses various features of each time point so that the data can participate in encoding and decoding without losing important information. The residual layer including convolution and LSTM further learns the dependence between time points and enables our model to carry out deeper training. A large number of experiments on three datasets show that the prediction results produced by our method are favorably comparable to the state-of-the-art.

## I. INTRODUCTION

Time series forecasting plays an important role in many fields. For example, the forecasting of sensor time series can detect the possible risks of equipment, which is of great significance to equipment safety and technological improvement. Weather changes are closely related to people's daily life. The analysis of weather changes over time can help people arrange their work and life in advance and prevent disasters caused by severe weather. In addition, time series forecasting also has tremendous applications in the domains of transportation, finance, and biological information.

Since the production of time series is usually influenced by various factors, many datasets in reality are multivariable time series. At the same time, due to the existence of uncertain factors, the time series data have certain randomness. If the model cannot correctly deal with these noises, there will be error accumulation in the training process, resulting in low prediction performance. In addition, time series data often changes nonlinearly, because there are many influencing factors, which makes it difficult to effectively learn and model time series trends. Over the past few decades, many researchers have proposed various forecasting methods based on the properties of time series. But for a long time, people have been stuck using pure mathematical means or shallow machine learning to process time series, such as autoregressive moving average model (ARIMA) [1], support vector machines (SVM) [2], and artificial neural networks (ANNs) [3], etc.

With the advent of the era of big data, the traditional mathematical equations and shallow learning models cannot effectively learn large-scale sequence features. Therefore, deep learning is introduced into time series forecasting. Recurrent neural network (RNN) is one of the most popular time series forecasting models. There are many variants based on RNN, such as long short-term memory (LSTM) [4], deep recurrent neural networks (DRNNs) [5], bidirectional recurrent neural networks (BRNNs) [6], etc. They can pay attention to the continuity of time and remember the information of the previous time, so as to learn the relevance before and after. However, the models based on RNN are sequential learning and cannot fully consider global information. Due to too many correlation parameters, it is easy to produce gradient disappearance and gradient explosion in the training process. Based on such problems, a parallel training model Transformer was proposed [7] and soon showed amazing expressiveness in the fields of natural language processing and computer vision. Then, various X-formers were proposed [8], which made a great contribution to enhancing attention and reducing parameter complexity. However, the classical Transformer is independent between batches during training, which ignores the coherence between batches. Moreover, because the transformer is trained in parallel, it cannot well learn the dependencies between adjacent time points. In addition, because almost all weight matrices in Transformer are linear, there will be a bottleneck in dealing with large-scale nonlinear time series. To solve the above problems, we propose a new Transformer based differential attention fusion model.

The main contributions of this paper are as follows:

- For the first time, a differential layer and differential preprocessing method are proposed to solve the problem of discontinuity between batches in the training process. We also propose a novel neighbor attention mechanism to make the model more sensitive to the differences of consecutive segments of time series data and learn nonlinearly dependent features more effectively.

- Moreover, we propose the sliding fusion mechanism to fuse various matrices generated by difference and neighbor attention mechanism, which can extract the

This work was supported by the Sichuan Science and Technology Program (NO.2021YFG0312) and the National Natural Science Foundation of China (NO.62176221).
∗ Shengdong Du is corresponding author.

time characteristics and relative importance of model dimensions, and make the model learn nonlinear features more effectively.

- Through experiments on two public datasets and a real private dataset, the proposed model shows good forecasting ability.

## II. RELATED WORK

Time series forecasting has always been the focus of people's attention. In the past few decades, people have developed a large number of time series forecasting models, but for a long time, people have studied how to predict the time series at the statistical level. Most methods first extract the relevant characteristics of the time series through mathematical means and then calculate the future value by formula, which have great limitations. Since the birth of neural networks, there have been many sequence prediction models based on deep learning. And a variety of models based on recurrent neural network (RNN) are widely used in the field of time series forecasting [9]. LSTM [4] is one of the representative models based on RNN. It uses three gating units to make it have long-term memory function. Subsequently, in order to solve the problem of the slow training speed of LSTM, a variant model GRU [10] is proposed to make time series forecasting more efficient. However, the early models based on RNN cannot learn the deep-seated dependence of time series. Therefore, attention mechanism is introduced into time series forecasting. An RNN model combined with an attention mechanism [11] can well capture the close relationship between input items, and the LSTM based on evolutionary attention [12] can mine the internal relationship of time series.

After the rise of the attention mechanism, Transformer based on the self-attention mechanism was proposed [7]. It was first applied in the field of natural language processing (NLP) and later proved to have strong ability in computer vision and other fields. After Transformer was proposed, many X-formers were proposed in order to enhance attention and solve memory problems [8]. Reformer replaces the dot-product attention mechanism with the method of locality-sensitive hashing [13], which reduces the complexity of the model. Informer proposes the ProbSparse self-attention mechanism, which makes the model have high prediction ability in long sequence time series [14].

In addition, some researchers proposed a difference long short-term memory network (DLSTM), which uses the differential method to extract dynamic temporal features [15]. Difference-attention mechanisms have also been proposed to focus on obvious changes in time series [16]. By comparing the differences between Transformer and various neural networks [17][18], people summarize the advantages and disadvantages of Transformer and design a variety of models combining Transformer and neural networks [19][20]. Moreover, with the increase of network depth, the prediction accuracy will be greatly reduced. In view of this phenomenon, [21] proposed the residual neural network.

The processing of sequences in Transformer is the masked multi-head attention, which obtains the relative importance of time points through the Q, K matrix, and covers up all the moments after the current time to ensure that the model can only learn before. However, this obviously cannot fully learn the time-varying features contained in the time series data. Moreover, in order to save memory, classical Transformer will set batches during training, but the training of each batch is independent, ignoring the impact of the previous batch on the next batch. In view of the in-depth analysis of the above problems, this paper proposes a new Transformer-based differential attention fusion model, which aims to deeply learn the changing characteristics of data and improve the shortcomings of the Transformer in time-dependent learning.

## III. METHODOLOGY

### A. Overview of the proposed model

In order to solve the above problems, we propose a differential attention fusion model based on Transformer. Fig. 1 shows the overall structure of the model. The model consists of an encoder and a decoder. In the encoder, the input data is divided into three adjacent parts in the differential layer. The purpose of the division is to introduce the difference between the two adjacent parts and ensure the consistency of the batches during training. The neighbor attention quickly learns the changing characteristics of the time series, and the sliding fusion mechanism fuses the obtained results in each dimension. In order to further learn temporal dependencies, we input the fused results into a one-dimensional convolutional neural network and LSTM. To avoid vanishing gradients, we add the output of the sliding fusion mechanism to the output of the neural networks to form a residual layer. In the decoder, it generates the prediction results by receiving the input of the original data and the output of the encoder.

### B. Differential Layer and Pos Encoding Layer

In order to fully learn the time advance features of the time series, we divide each training input $X_{1:N} \in \mathbb{R}^{N \times d_{input}}$ into three parts. The result is $\{x^{(j)}|X_{j:N-3+j}, j \in \{1,2,3\}\}$. Take $x^{(2)}$ as the training center, the forward part is $x^{(1)}$ and the backward part is $x^{(3)}$. In this way, when training the next batch, the previous training center will become the forward part of the current training, and the backward part of the previous training will become the training center. In order to better learn the correlation between various parts, we make a difference to $x^{(j)}$ and carry out input embedding and positional encoding respectively. The positional encoding formula used in this paper is:

$$PE_{(p,2i)} = \sin(pos/10000^{2i/d_{model}}) \quad (1)$$

$$PE_{(p,2i+1)} = \cos(pos/10000^{2i/d_{model}}) \quad (2)$$

where $p$ is the position and $i$ is the dimension.

Then we add the positional encoding matrices to the input embedding results. These two steps are completed together in pos encoding layer in Fig. 1. The complete process of differential layer, input embedding, and pos encoding layer is as follows

$$h_{F,C,B} = x^{(j)}W_j + PE(x^{(j)}W_j), j \in \{1,2,3\} \quad (3)$$

$$D_F = (x^{(2)} - x^{(1)})W_F + PE((x^{(2)} - x^{(1)})W_F) \quad (4)$$

$$D_B = (x^{(2)} - x^{(3)})W_B + PE((x^{(2)} - x^{(3)})W_B) \quad (5)$$

Among them, $h_{F,C,B}$ represents the forward, center, and backward coding matrices when $j$ is 1, 2, and 3, respectively. $h_{F,C,B}, D_F, D_B \in \mathbb{R}^{(N-2) \times d_{model}}$. Note $n = N - 2$. Since $D_F$ and $D_B$ have the same structure, we will use $D_{F,B}$ to represent this type of matrix uniformly below.

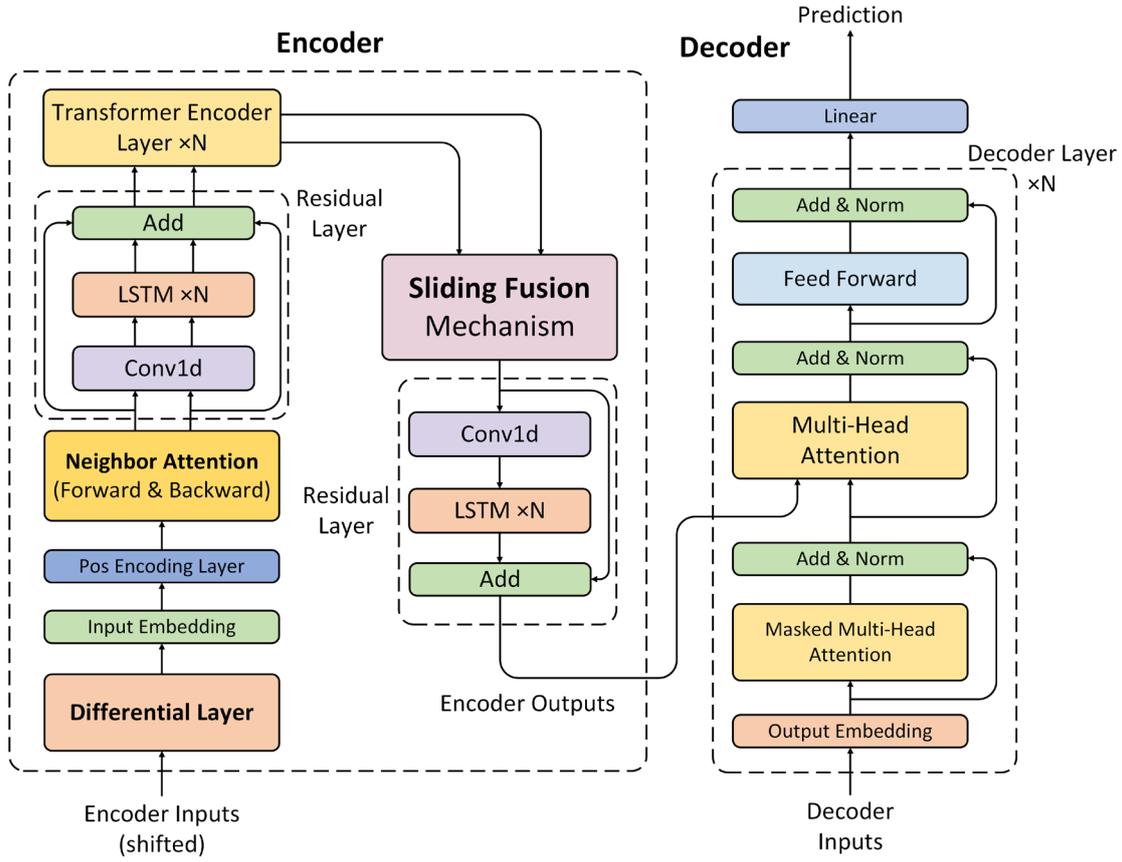

Fig. 1. The architecture of the proposed model

## C. Neighbor Attention and Sliding Fusion Mechanism

Before entering the encoder, the neighbor attention mechanism is introduced, which includes the forward and the backward attention mechanism, both of which include the multi-head attention and the sliding fusion mechanism, as shown in Fig. 2(b). In order to get the value $A_{F,B} \in \mathbb{R}^{n \times d_{model}}$ of the training center's attention to the forward and backward parts, we first use the method of linear mapping to obtain $Q_{F,B} \in \mathbb{R}^{n \times d_{attn}}$ from $h_C$, and generate $K_{F,B} \in \mathbb{R}^{n \times d_{attn}}$ from $h_F$ and $h_B$, then get $V_{F,B} \in \mathbb{R}^{n \times d_{model}}$ from $h_C$. Next, we divide $Q_{F,B}K_{F,B}^T \in \mathbb{R}^{n \times n}$ by $\sqrt{d_{attn}}$ and use the softmax function to obtain the relative attention intensity between time points. The complete formula is as follows:

$$A_{F,B} = \text{softmax}\left(\frac{Q_{F,B}K_{F,B}^T}{\sqrt{d_{attn}}}\right)V_{F,B} \quad (6)$$

The multi-head attention mechanism in the classical Transformer helps the model to focus on the information of different representation subspaces at different locations [7]. This paper also adopts this mechanism, as shown in Fig. 2(a). The specific formula is as follows:

$$MultiHead(Q_{F,B}, K_{F,B}, V_{F,B}) = Concat(head_1, \dots, head_p)W^O$$

$$\text{where } head_i = A_{F,B}^{(i)} \quad (7)$$

where $W^O \in \mathbb{R}^{pd_{model} \times d_{model}}$.

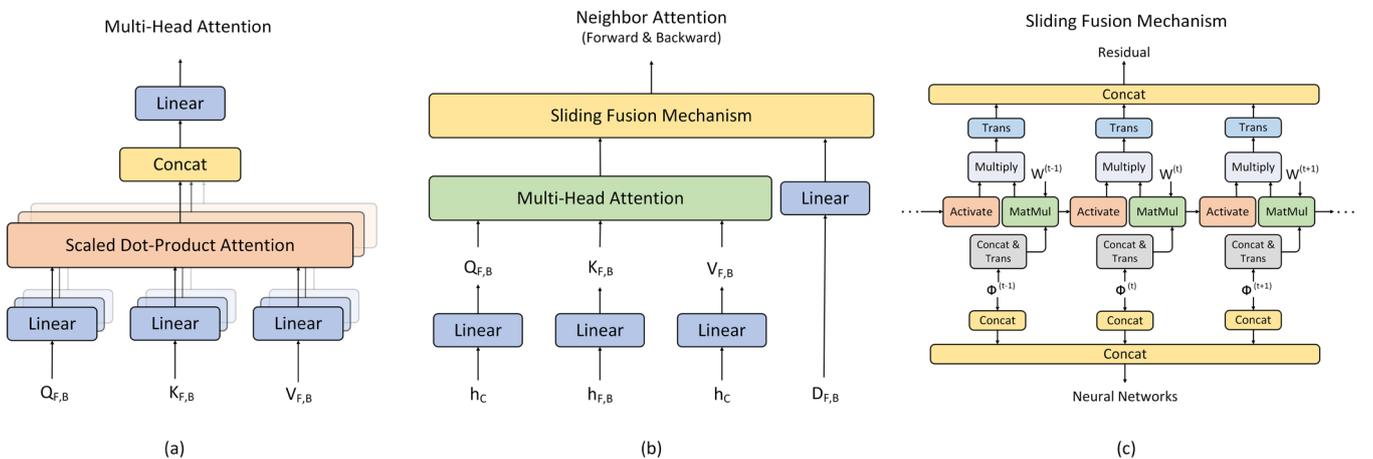

Fig. 2. (a)Multi-Head Attention, (b)Neighbor Attention, (c)Sliding Fusion Mechanism

The above three types of matrices $D_{F,B}, h_C, A_{F,B}$ are obtained. Since the influence of difference results, training center, and attention value on prediction at different time points is unknown, we select the three types of matrices and splice them at each same time point to obtain matrix $c_{F,B}^{(t)} \in \mathbb{R}^{3 \times d_{mdoel}}$, transpose them and then multiply them by matrix $W_{F,B}^{(t)} \in \mathbb{R}^{3 \times 1}$ to obtain the weighted results. Considering the influence of the previous time on the current time and adding nonlinear factors, we use the activation function σ to extract the relative importance of each dimension from the weighted result of the previous time, and then multiply it with the corresponding position of the weighted result of the current time to obtain the final result $f_{F,B}^{(t)} \in \mathbb{R}^{d_{model} \times 1}$. The matrix $f_{F,B}^{(t)}$ obtained at each time $t$ is transposed and spliced to obtain $e_{F,B} \in \mathbb{R}^{n \times d_{model}}$. The process is shown in Fig. 2(c), and the general calculation formula is as follows:

$$c^{(t)} = Concat(\Phi^{(t)}) \tag{8}$$

$$f^{(t)} = (c^{(t)})^T W^{(t)} \odot s^{(t-1)} \tag{9}$$

$$s^{(t)} = \sigma((c^{(t-1)})^T W^{(t-1)}) \tag{10}$$

$$e = Concat(f^{(t)})^T \tag{11}$$

where $\Phi^{(t)}$ represents the set of various matrices at time $t$.

*D. Residual Layer*

For the splicing matrix $c_{F,B}$ obtained above, we hope to extract the features of each time point, so as to make better use of the time series prediction model for training. According to the feature that $c_{F,B}$ has three dimensions at each time point, our model introduces a one-dimensional convolution neural network to process it. The convolution kernel size is 3 and the step size is 3. The purpose is to extract the features at each time point. The number of convolution kernels represents the feature dimension at each time point after extraction, which can be set freely according to different scenes.

Our model combines multiple LSTM layers with the above convolution layer and uses its long-term memory function to deeply extract the time dependence to make up for the loss of time information caused by the parallel training of the original Transformer. We also add $e_{F,B}$ as the residual to the final output of LSTM as the input of the encoder. The mechanism is shown in Fig. 1. The output result of the encoder is $o_{F,B} \in \mathbb{R}^{n \times d_{model}}$, which is processed by the attention mechanism and neural networks mentioned above to obtain the input of the decoder. If you simply put the output results into two different decoders and combine them after decoding, there will be a mismatch. Therefore, we introduce the sliding fusion mechanism and residual layer at the connection between encoder and decoder, which can also make aggregation analysis on the forward and backward dimensions to improve the prediction accuracy of the model.

IV. EXPERIMENTS AND RESULTS ANALYSIS

In this section, we conduct experiments using real time series datasets to analyze and evaluate the proposed model. Compared with the shallow learning models, recurrent neural networks, and Transformer-based models, the prediction performance and effectiveness of the proposed model are verified. In addition, we also conduct the ablation analysis of the proposed model on three datasets to compare the contribution of different modules to the prediction accuracy.

*A. Datasets and Experimental Environment*

In order to evaluate the performance of the proposed model, we use two public datasets and a real private dataset. The public dataset is the American climatological dataset, Geo-Magnetic field and WLAN dataset. The private dataset is the shield data of the Qinghuayuan Tunnel in Beijing, China. The details of these three datasets are as follows.

**Weather**(local climatological dataset): This dataset contains the local climate data of nearly 1600 locations in the United States from 2010 to 2013, and the data is collected every hour. It contains the prediction target Wet Bulb Celsius and 11 related parameters. The dataset can be downloaded from the data website of the national environmental information center[1].

**GM-WLAN**(Geo-Magnetic field and WLAN dataset): This dataset carries out indoor positioning through personal devices such as smartwatches and smartphones with built-in sensors, including acceleration, magnetic field, and azimuth in all directions. We predict the acceleration in the X-direction. This dataset was shared in the UCI Machine Learning Repository[2].

**Tunnel**(Qinghuayuan Tunnel shield dataset): This dataset is a multivariable time series dataset, which contains the ring number of shield machines and 20 parameters used in the experiment. The data is advanced according to the number of rings. Our experiment predicts two parameters, namely torque, and rotational speed(In the following text, we use speed instead of rotational speed).

Then we describe the hardware and software of the experiment, as well as the configuration of related parameters. The open-source deep learning framework Pytorch is used to build Transformer based models, and Karas is used to build RNN based models. All experiments were conducted on the computer side, configured with Intel (R) Xeon (R) CPU e5-2687 V3 @ 2.50ghz 2.50 GHz (2 processors).

*B. Model Setting*

For the tunnel dataset, we use the first 360 pieces of data as the training set and the next 155 pieces of data as the test set. For the weather dataset, we use the data of January (31 days) as the training set to predict the first 15 days of February. For the Geo-Magnetic field and WLAN dataset, we select the 100th to 460th data of the first smartphone as the training set and the subsequent 155 pieces of data as the test set. Since the data generated by the gyroscope are all 0, we abandon the three attributes generated by the gyroscope during training and testing. In experiments, our model is compared with RNN, LSTM, GRU, classical Transformer, and Informer.

In order to ensure that every piece of data in the training set can be used, that is to say, each piece of data should be included in the training center, we shift the original data forward and backward by one time point when inputting, which allows our model to run completely without affecting the final prediction. During the test, we can also set the batch.

---

[1] https://www.ncei.noaa.gov/data/local-climatological-data/

[2] https://archive.ics.uci.edu/ml/datasets/Geo-Magnetic+field+and+WLAN+dataset+for+indoor+localisation+from+wristband+and+smartphone

The batch size here can be different from the batch size set during the training. Finally, we only select the predicted value of each batch at the first time point of the training center as the final data. The results predicted by the latter batch can cover all the results predicted by the previous batch except the first data. In this way, each batch only leaves a prediction value at the first time point of the training center, which improves the accuracy of the model prediction.

We select the mean square error (MSE) as the loss function of our model. The Sigmoid function is chosen as the activation function in the sliding fusion mechanism. We normalize the data to [0,1] by the min-max method and adopt the decreasing learning rate. The decreasing mechanism is as follows:

$$lrate_{(epoch)} = lrate_{(epoch-1)} \times 0.95^{epoch} \quad (12)$$

Among them, the epoch represents the number of cyclic training of the model, and the initial learning rate is set to 0.0005. In the one-dimensional convolution neural network before inputting the encoder, we set the convolution kernel size as $\mathbb{R}^{3 \times d_{model}}$, the step size as 3, the number of convolution kernels as 16. Set two long short-term memory networks (LSTM) after convolution. Set the input dimension of the first LSTM as 16, the output dimension as 32, the input dimension of the second LSTM as 32, the output dimension as $d_{model}$, and the dropout as 0.5. In the one-dimensional convolution neural network at the connection of encoder and decoder, we set the convolution kernel size as $\mathbb{R}^{2 \times d_{model}}$, the step size as 2, and the number of convolution kernels as 16. After convolution, we still set two LSTM, and their parameter settings are the same as those after the neighbor attention mechanism. Adam is selected as the optimizer of model training. The batch size of training and testing is 20. Finally, the mean absolute error (MAE) and root mean squared error (RMSE) are selected as the evaluation indicators of the model.

*C. Results Analysis*

Table I shows the comparison of prediction performance of shallow learning model ARIMA, classical deep learning model (RNN, LSTM, GRU, Transformer, Informer), and our model on three time series datasets. The experimental results show that our model is better than other models. Compared with LSTM, our model reduces RMSE by 81.3%, 37.1%, 65.0%, and 45.1% on Tunnel(torque), Tunnel (speed), Weather, and GM-WLAN datasets respectively, which shows that our method has greatly improved the prediction ability of RNN based model. Compared with classical Transformer, our model reduces RMSE by 45.2% (Tunnel (torque)), 54.9% (Tunnel (speed)), 76.9% (Weather), 38.4% (GM-WLAN), which shows that the attention to difference and the residual layer integrated with convolution neural network and LSTM can better learn the changes and time dependence that may affect the global situation. Compared with Informer, our method reduces RMSE by 46.7% (Weather) and 62.1% (GM-WLAN), which shows that our model can quickly learn the changes affecting the overall situation in a short time compared with Informer suitable for long-distance prediction.

Table II shows the forecasting results of the proposed model without the differential layer, neighbor attention, and sliding fusion mechanism. By comparison with Table I, it can be seen that the attention to difference can reduce RMSE by 36.7% (Tunnel (torque)), 49.2% (Tunnel(speed)), 47.5% (Weather), 26.9% (GM-WLAN), which indicates that the attention mechanism on difference can greatly improve the prediction performance. Table III shows the prediction results of removing the residual layer. By comparison with Table I, it can be seen that after incorporating convolution neural network and LSTM, our model can further reduce RMSE by 10.5% (Tunnel (torque)), 33.3% (Tunnel (speed)), 50.6% (Weather), 41.5% (GM-WLAN), which indicates that the neural network can make up for the deficiencies of the Transformer architecture and further capture the time dependence. Comparing Table II and Table III with the Transformer in Table I, we find that the model incorporating differential attention reduces RMSE by 38.8% (Tunnel(torque)), 32.4% (Tunnel(speed)), and 53.3% (Weather), 30.8% (GM-WLAN), the model incorporating convolution and LSTM reduced RMSE by 13.3% (tunnel (torque)), 11.3% (tunnel (speed)), 56.0% (weather), 15.7% (GM-WLAN), the effect of the neural network on RMSE is significantly smaller than that on difference attention, which indicates that difference-based attention plays a major role in improving the prediction performance of the proposed model, and the residual layer incorporating convolution and LSTM can help the model learn temporal features and dependencies better, and give our model better training performance.

TABLE I. MULTIVARIATE TIME SERIES FORECASTING RESULTS ON THREE DATASETS

| Models | Datasets | | | | | | | |
|---|---|---|---|---|---|---|---|---|
| | Tunnel(torque) | | Tunnel(speed) | | Weather | | GM-WLAN | |
| | MAE | RMSE | MAE | RMSE | MAE | RMSE | MAE | RMSE |
| ARIMA | 3.577 | 4.055 | 0.098 | 0.108 | 2.393 | 2.954 | 0.284 | 0.479 |
| RNN | 3.995 | 4.605 | 0.044 | 0.055 | 0.651 | 0.883 | 0.713 | 1.017 |
| LSTM | 3.685 | 4.238 | 0.039 | 0.055 | 0.645 | 0.863 | 0.243 | 0.386 |
| GRU | 3.675 | 4.022 | 0.028 | 0.045 | 0.641 | 0.885 | 0.327 | 0.484 |
| Transformer | 1.290 | 1.445 | 0.065 | 0.071 | 1.083 | 1.308 | 0.249 | 0.344 |
| Informer | 0.891 | 1.219 | 0.379 | 0.485 | 0.389 | 0.567 | 0.348 | 0.559 |
| **Ours** | **0.643** | **0.792** | **0.025** | **0.032** | **0.236** | **0.302** | **0.128** | **0.212** |

Note: For the same dataset, different attributes of prediction are indicated in the bracket.

TABLE II.  MODEL ABLATION ANALYSIS (WITHOUT DIFFERENTIAL LAYER, NEIGHBOR ATTENTION, AND SLIDING FUSION MECHANISM)

| Datasets | Metrics | |
|---|---|---|
| | MAE | RMSE |
| **Tunnel(torque)** | 1.146 | 1.252 |
| **Tunnel(speed)** | 0.053 | 0.063 |
| **Weather** | 0.531 | 0.575 |
| **GM-WLAN** | 0.236 | 0.290 |

TABLE III.  MODEL ABLATION ANALYSIS (WITHOUT RESIDUAL LAYER)

| Datasets | Metrics | |
|---|---|---|
| | MAE | RMSE |
| **Tunnel(torque)** | 0.708 | 0.885 |
| **Tunnel(speed)** | 0.039 | 0.048 |
| **Weather** | 0.451 | 0.611 |
| **GM-WLAN** | 0.180 | 0.238 |

With the advancement of the shield machine, the properties of the soil layer will also continue to change, so the parameters of the shield machine will obtain a certain trend. It can be seen from Fig. 3 that with the advancement of the number of rings (the x-axis represents the ring number), the depth of the tunnel is also increasing, and the torque of the shield machine is in an upward trend. Our model can learn this trend well, while the original transformer does not pay enough attention to the difference of data, so it is not good to learn the trend here. The model based on RNN cannot learn the global information well, so it is not sensitive to the trend here, and the predicted results still fluctuate within a certain range.

Due to the different behaviors of people at different time points, the indoor positioning data generated by wearable devices have great uncertainty. Human perception devices are very sensitive to behavior mutation. The mutation is not caused by random factors, but human behavior, which will make the data predictable. As can be seen from Fig. 4, in a period of time, there are many sudden changes in the acceleration in the X-direction, the biggest sudden change is at the middle time of our prediction, and there are many small sudden changes in the rest. Our model has a good fitting effect on the real value. Transformer and RNN based models do not grasp the mutation well and are vulnerable to the negative impact of the mutation at other time points.

In summary, the prediction results of our proposed model are in good agreement with the real situation and can learn the changing characteristics of time series data in a short time, which is also better than Transformer and RNN based models. This shows that the differential attention fusion model based on Transformer can effectively capture the temporal dependence of time series data and gain insight into its internal development patterns.

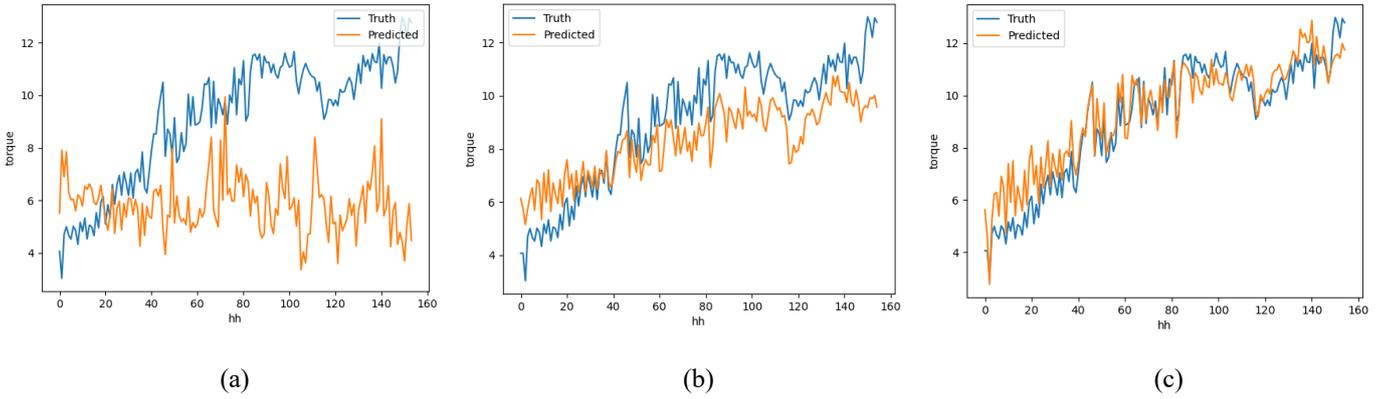

Fig. 3. (a)LSTM, (b)Transformer, (c)Our model predicts the torque on the Qinghuayuan Tunnel shield dataset. The predicted length is 155, blue represents the real value and orange represents the predicted value

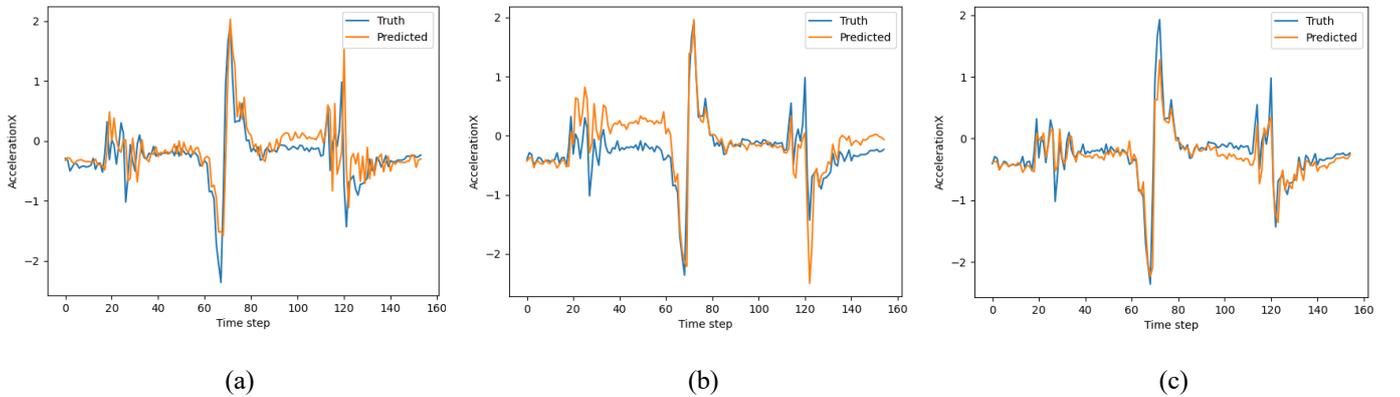

Fig. 4. (a)LSTM, (b)Transformer, (c)Our model predicts the acceleration x on the Geo-Magnetic field and WLAN dataset. The prediction length is 155, blue represents the real value and orange represents the predicted value.

## V. CONCLUSIONS

In this paper, we propose a differential input form for time series, which makes the model more temporally coherent during training, and also design the neighbor attention and the sliding fusion mechanism based on the differential form, which extends the original Transformer attention mechanism. With the addition of the residual layer that incorporates neural networks, the proposed model can learn the time dependence at a deeper level, and enhance the forecasting ability. Experiments on real data demonstrate that our model is able to make efficient predictions on different time series datasets generated on a variety of occasions. The model pays special attention to the difference of data and is very sensitive to the changing characteristics of time series, which provides a useful reference for the maintenance of industrial equipment and the prediction of seasonal phenomena in real life. In future research, the specific form of the difference can be further improved, and the attention mechanism for the difference can be designed to save more memory.

### ACKNOWLEDGMENT

This work was supported by the Sichuan Science and Technology Program (NO.2021YFG0312) and the National Natural Science Foundation of China (NO.62176221).